\documentclass[10pt,twocolumn,letterpaper]{article} 

\usepackage{iccv}
\usepackage{times}
\usepackage{epsfig}
\usepackage{graphicx}
\usepackage{amsmath}
\usepackage{amssymb}
\usepackage{float}
\usepackage{multirow}
\usepackage{authblk}

\usepackage[pagebackref=true,breaklinks=true,letterpaper=true,colorlinks,bookmarks=false]{hyperref}

\iccvfinalcopy 

\def\iccvPaperID{15} 

\ificcvfinal\fi
\begin{document}

\title{Learning to Detect Violent Videos using Convolutional Long Short-Term Memory}

\author[1,2]{Swathikiran Sudhakaran}
\author[2]{Oswald Lanz}
\affil[1]{University of Trento, Trento, Italy}
\affil[2]{Fondazione Bruno Kessler, Trento, Italy}
\affil[ ]{\tt\small {\{sudhakaran,lanz\}@fbk.eu}}

\maketitle
 \thispagestyle{empty}
\begin{abstract}
Developing a technique for the automatic analysis of surveillance videos in order to identify the presence of violence is of broad interest. In this work, we propose a deep neural network for the purpose of recognizing violent videos. A convolutional neural network is used to extract frame level features from a video. The frame level features are then aggregated using a variant of the long short term memory that uses convolutional gates. The convolutional neural network along with the convolutional long short term memory is capable of capturing localized spatio-temporal features which enables the analysis of local motion taking place in the video. We also propose to use adjacent frame differences as the input to the model thereby forcing it to encode the changes occurring in the video. The performance of the proposed feature extraction pipeline is evaluated on three standard benchmark datasets in terms of recognition accuracy. Comparison of the results obtained with the state of the art techniques revealed the promising capability of the proposed method in recognizing violent videos. 
\end{abstract}

\section{Introduction}

Nowadays, the amount of public violence has increased dramatically. This can be a terror attack involving one or a number of persons wielding guns to a knife attack by a single person. This has resulted in the ubiquitous usage of surveillance cameras. This has helped authorities in identifying violent attacks and take the necessary steps in order to minimize the disastrous effects. But almost all the systems nowadays require manual human inspection of these videos for identifying such scenarios, which is practically infeasible and inefficient. It is in this context that the proposed study becomes relevant. Having such a practical system that can automatically monitor surveillance videos and identify the violent behavior of humans will be of immense help and assistance to the law and order establishment. In this work, we will be considering aggressive human behavior as violence rather than the presence of blood or fire.

The development of several deep learning techniques, brought about by the availability of large datasets and computational resources, has resulted in a landmark change in the computer vision community. Several techniques with improved performance for addressing problems such as object detection, recognition, tracking, action recognition, caption generation, etc. have been developed as a result. However, despite the recent developments in deep learning, very few deep learning based techniques have been proposed to tackle the problem of violence detection from videos. Almost all the existing techniques rely on hand-crafted features for generating visual representations of videos. The most important advantage of deep learning techniques compared to the traditional hand-crafted feature based techniques is the ability of the former to achieve a high degree of generalization. Thus they are able to handle unseen data in a more effective way compared to hand-crafted features. Moreover, no prior information about the data is required in the case of a deep neural network and they can be inputted with raw pixel values without much complex pre-processing. Also, deep learning techniques are not application specific unlike the hand-crafted feature based methods since a deep neural network model can be easily applied for a different task without any significant changes to the architecture. Owing to these reasons, we choose to develop a deep neural network for performing violent video recognition.

Our contributions can be summarized as follows:
\begin{itemize}
	\item We develop an end-to-end trainable deep neural network model for performing violent video classification
	\item We show that a recurrent neural network capable of encoding localized spatio-temporal changes generates a better representation, with less number of parameters, for detecting the presence of violence in a video
	\item We show that a deep neural network trained on the frame difference performs better than a model trained on raw frames
	\item We experimentally validate the effectiveness of the proposed method using three widely used benchmarks for violent video classification
\end{itemize}

The rest of the document is organized as follows. Section 2 discusses some of the relevant techniques for performing violent video recognition followed by a detailed explanation of the proposed deep neural network model in Section 3. The details about the various experiments conducted as part of this research are given in Section 4 and the document is concluded in Section 5.

\section{Related Works}
\begin{figure*}[t]
	\begin{center}
		\includegraphics[scale=0.26]{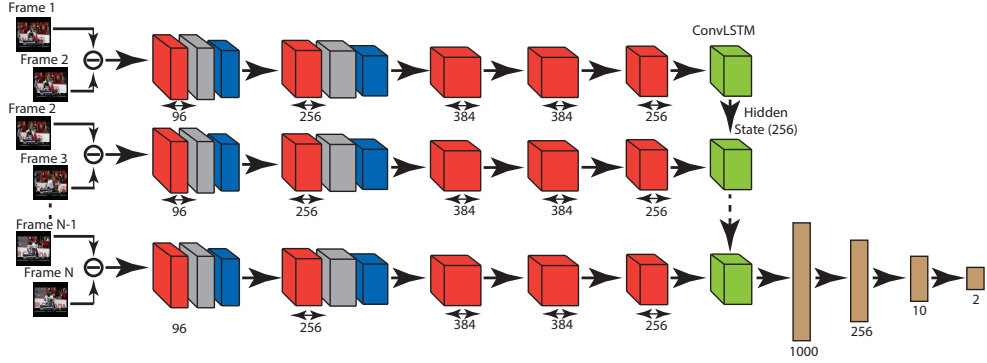}	\end{center}
	\caption{Block diagram of the proposed model. The model consists of alternating convolutional (red), normalization (grey) and pooling (blue) layers. The hidden state of the ConvLSTM (green) at the final time step is used for classification. The fully-connected layers are shown in brown colour.}
	\label{fig:block_dia}
\end{figure*}

Several techniques have been proposed by researchers for addressing the problem of violence detection from videos. These include methods that use the visual content \cite{nievas2011violence,bilinski2016human}, audio content \cite{pfeiffer1997automatic,giannakopoulos2007multi} or both \cite{zajdel2007cassandra,acar2016breaking}. In this section, we will be concentrating on methods that use the visual cues alone since it is more related to the proposed approach and moreover audio data is generally unavailable with surveillance videos. All the existing techniques can be divided into two classes depending on the underlying idea\\ 1. Inter-frame changes: Frames containing violence undergo massive variations because of fast motion due to fights \cite{vasconcelos1997towards,clarin2005dove,chen2011violence,deniz2014fast}\\
2. Local motion in videos: The motion change patterns taking place in the video is analyzed \cite{datta2002person,chen2008recognition,de2010violence,nievas2011violence,Hassneretal:SISM12,xu2014violent,mohammadi2015violence,rota2015real,gracia2015fast,bilinski2016human,gao2016violence,zhang2017discriminative}

Vasconcelos and Lippman \cite{vasconcelos1997towards} used the tangent distance between adjacent frames for detecting the inter-frame variations. Clarin et al. improves this method in \cite{clarin2005dove} by finding the regions with skin and blood and analyzing these regions for fast motion. Chen et al. \cite{chen2011violence} uses the motion vector encoded in the MPEG-1 video stream for detecting frames with high motion content and then detects the presence of blood for classifying the video as violent. Deniz et al. \cite{deniz2014fast} proposes to use the acceleration estimate computed from the power spectrum of adjacent frames as an indicator of fast motion between successive frames.

Motion trajectory information and the orientation of limbs of the persons present in the scene is proposed as a measure for detecting violence by Datta et al. \cite{datta2002person}. Several other methods follow the techniques used in action recognition, i.e., to identify spatio-temporal interest points and extract features from these points. These include Harris corner detector \cite{chen2008recognition}, Space-time interest points (STIP) \cite{de2010violence}, motion scale-invariant feature transform (MoSIFT) \cite{nievas2011violence,xu2014violent}. Hassner et al. \cite{Hassneretal:SISM12} introduces a new feature descriptor called violent flows (ViF), which is the flow magnitude over time of the optical flow between adjacent frames, for detecting violent videos. This method is improved by Gao et al. \cite{gao2016violence} by incorporating the orientation of the violent flow features resulting in oriented violent flows (OViF) features. Substantial derivative, a concept in fluid dynamics, is proposed by Mohammadi et al. \cite{mohammadi2015violence} as a discriminative feature for detecting violent videos. Gracia et al. \cite{gracia2015fast} proposes to use the blob features, obtained by subtracting adjacent frames, as the feature descriptor. The improved dense trajectory features commonly used in action recognition is used as a feature vector by Bilinski et al. in \cite{bilinski2016human}. They also propose an improved Fisher encoding technique that can encode spatio-temporal position of features in a video. Zhang et al. \cite{zhang2017discriminative} proposes to use a modified version of motion Weber local descriptor (MoIWLD) followed by sparse representation as the feature descriptor.

The hand-crafted feature based techniques used methods such as bag of words, histogram, improved Fisher encoding, etc. for aggregating the features across the frames. Recently various models using long short term memory (LSTM) RNNs \cite{hochreiter1997long} have been developed for addressing problems involving sequences such as machine translation \cite{sutskever2014sequence}, speech recognition \cite{graves2013hybrid}, caption generation \cite{xu2015show,venugopalan2015sequence} and video action recognition \cite{donahue2015long,srivastava2015unsupervised}. The LSTM was introduced in 1997 to combat the effect of vanishing gradient problem which was plaguing the deep learning community. The LSTM incorporates a memory unit which contains information about the inputs the LSTM unit has seen and is regulated using a number of fully-connected gates. The same idea of using LSTM for feature aggregation is proposed by Dong et al. in \cite{dong2016multi} for violence detection. The method consisted of extracting features using a convolutional neural network from raw pixels, optical flow images and acceleration flow maps followed by LSTM based encoding and a late fusion.

Recently, Xingjian et al. \cite{xingjian2015convolutional} replaced the fully-connected gate layers of the LSTM with convolutional layers and used this improved model for predicting precipitation nowcasting from radar images with improved performance. This newer model of the LSTM is named as convolutional LSTM (convLSTM). Later, it has been used for predicting optical flow images from videos \cite{PatrauceanHC16} and for anomaly detection in videos \cite{medel2016anomaly}. By replacing the fully-connected layers in the LSTM with convolutional layers, the convLSTM model is capable of encoding spatio-temporal information in its memory cell.

\section{Proposed method}

The goal of the proposed study was to develop an end-to-end trainable deep neural network model for classifying videos in to violent and non-violent ones. The block diagram of the proposed model is illustrated in figure \ref{fig:block_dia}. The network consists of a series of convolutional layers followed by max pooling operations for extracting discriminant features and convolutional long short memory (convLSTM) for encoding the frame level changes, that characterizes violent scenes, existing in the video.

\subsection{ConvLSTM}
Videos are sequences of images. For a system to identify if a fight is taking place between the humans present in the video, it should be capable of identifying the locations of the humans and understand how the motion of the said humans are changing with time. Convolutional neural networks (CNN) are capable of generating a good representation of each video frame. For encoding the temporal changes a recurrent neural network (RNN) is required. Since we are interested in changes in both the spatial and temporal dimensions, convLSTM will be a suitable option. Compared to LSTM, the convLSTM will be able to encode the spatial and temporal changes using the convolutional gates present in them. This will result in generating a better representation of the video under analysis.
 The equations of the convLSTM model are given in equations 1-6.
\begin{eqnarray}
i_t = \sigma(w_x^i* I_t + w_h^i*h_{t-1} + b^i)\\
f_t = \sigma(w_x^f* I_t + w_h^f*h_{t-1} + b^f)\\
\tilde{c}_t = tanh(w_x^{\tilde{c}}*I_t + w_h^{\tilde{c}}*h_{t-1} + b^{\tilde{c}})\\
c_t = \tilde{c}_t\odot i_t + c_{t-1}\odot f_t\\
o_t = \sigma(w_x^o * I_t + w_h^o*h_{t-1} + b^o)\\
h_t = o_t\odot tanh(c_t)
\end{eqnarray}

In the above equations, `*' represents convolution operation and `$\odot$' represents the Hadamard product. The hidden state $h_t$, the memory cell $c_t$ and the gate activations $i_t$, $f_t$ and $o_t$ are all 3D tensors in the case of convLSTM.

For a system to identify a video as violent or non-violent, it should be capable of encoding localized spatial features and the manner in which they change with time. Hand-crafted features are capable of achieving this with the downside of having increased computational complexity. CNNs are capable of generating discriminant spatial features but existing methods use the features extracted from the fully-connected layers for temporal encoding using LSTM. The output of the fully-connected layers represents a global descriptor of the whole image. Thus the existing methods fail to encode the localized spatial changes. As a result, they resort to methods involving addition of more streams of data such as optical flow images \cite{dong2016multi} which results in increased computational complexity. It is in this context that the use of convLSTM becomes relevant as it is capable of encoding the convolutional features of the CNN. Also, the convolutional gates present in the convLSTM is trained to encode the temporal changes of local regions. In this way, the whole network is capable of encoding localized spatio-temporal features.

\subsection{Network Architecture}
Figure \ref{fig:block_dia} illustrates the architecture of the network used for identifying violent videos. The convolutional layers are trained to extract hierarchical features from the video frames and are then aggregated using the convLSTM layer. The network functions as follows: The frames of the video under consideration are applied sequentially to the model. Once all the frames are applied, the hidden state of the convLSTM layer in this final time step contains the representation of the input video frames applied. This video representation, in the hidden state of the convLSTM, is then applied to a series of fully-connected layers for classification.

In the proposed model, we used the AlexNet model \cite{krizhevsky2012imagenet} pre-trained on the ImageNet database as the CNN model for extracting frame level features. Several studies have found out that networks trained on the ImageNet database is capable of having better generalization and results in improved performance for tasks such as action recognition \cite{simonyan2014two} \cite{misra2016shuffle}. In the convLSTM, we used 256 filters in all the gates with a filter size of $3\times 3$ and stride 1. Thus the hidden state of the convLSTM consists of 256 feature maps. A batch normalization layer is added before the first fully-connected layer. Rectified linear unit (ReLU) non-linear activation is applied after each of the convolutional and fully-connected layers.

In the network, instead of applying the input frames as such, the difference between adjacent frames are given as input. In this way, the network is forced to model the changes taking place in adjacent frames rather than the frames itself. This is inspired by the technique proposed by Simonyan and Zisserman in \cite{simonyan2014two} to use optical flow images as input to a neural network for action recognition. The difference image can be considered as a crude and approximate version of optical flow images. So in the proposed method, the difference between adjacent video frames are applied as input to the network. As a result, the computational complexity involved in the optical flow image generation is avoided. The network is trained to minimize the binary cross entropy loss.
\vskip -6mm
\section{Experiments and Results}
\begin{table}[t]
\centering
\caption{Classification accuracy obtained with the hockey fight dataset for different models}
\begin{tabular}{l c}
	\hline
	\textbf{Input} & \textbf{Classification Accuracy}\\ \hline \hline
	Video Frames & \multirow{2}{*}{94.1$\pm$2.9\%}\\
	(random initialization) & \\
	Video Frames & \multirow{2}{*}{96$\pm$0.35\%}\\
	(ImageNet pre-trained) & \\
	Difference of Video Frames & \multirow{2}{*}{95.5$\pm$0.5\%}\\
	(random initialization) & \\
	Difference of Video Frames & \multirow{2}{*}{97.1$\pm$0.55\%}\\
	(ImageNet pre-trained) & \\ \hline
\end{tabular}
\label{table2}
\end{table}

\begin{table*}[t]
	\centering
	\caption{Comparison of classification results}
	\begin{tabular}{l c c c}
		\hline
		\textbf{Method} & \textbf{Hockey Dataset} & \textbf{Movies Dataset} & \textbf{Violent-Flows Dataset}\\ \hline \hline
		MoSIFT+HIK\cite{nievas2011violence} & 90.9\% & 89.5\% & -\\ 
		ViF\cite{Hassneretal:SISM12} & 82.9$\pm$0.14\% & - & 81.3$\pm$0.21\%\\
		MoSIFT+KDE+Sparse Coding\cite{xu2014violent} & 94.3$\pm$1.68\% & - & 89.05$\pm$3.26\%\\
		Deniz et al.\cite{deniz2014fast} & 90.1$\pm$0\% & 98.0$\pm$0.22\% & -\\
		Gracia et al.\cite{gracia2015fast} & 82.4$\pm$0.4\% & 97.8$\pm$0.4\% & -\\
		Substantial Derivative\cite{mohammadi2015violence} & - & 96.89$\pm$0.21\% & 85.43$\pm$0.21\%\\
		Bilinski et al.\cite{bilinski2016human} & 93.4 & 99 & \textbf{96.4}\\
		MoIWLD\cite{zhang2017discriminative} & 96.8$\pm$1.04\% & - & 93.19$\pm$0.12\%\\
		ViF+OViF\cite{gao2016violence} & 87.5$\pm$1.7\% & - & 88$\pm$2.45\%\\
		Three streams + LSTM\cite{dong2016multi} & 93.9 & - & -\\
			\textbf{Proposed} & \textbf{97.1$\pm$0.55\%} & \textbf{100$\pm$0\%} & 94.57$\pm$2.34\% \\ \hline
	\end{tabular}
	\label{table1}
\end{table*}
\vskip -1mm
To evaluate the effectiveness of the proposed approach in classifying violent videos, three benchmark datasets are used and the classification accuracy is reported.

\subsection{Experimental Settings}
The network is implemented using the Torch library. From each video, $N$ number of frames equally spaced in time are extracted and resized to a dimension of $256\times 256$ for training. This is to avoid the redundant computations involved in processing all the frames, since adjacent frames contain overlapping information. The number of frames selected is based on the average duration of the videos present in each dataset. The network is trained using RMSprop algorithm with a learning rate of $10^{-4}$ and a batch size of 16. The model weights are initialized using Xavier algorithm. Since the number of videos present in the datasets are limited, data augmentation techniques such as random cropping and horizontal flipping are used during training stage. During each training iteration, a portion of the frame of size $224\times 224$ is cropped, from the four corners or from the center, and is randomly flipped before applying to the network. Note that the same augmentation technique is followed for all the frames present in a video. The network is run for 7500 iterations during the training stage. In the evaluation stage, the video frames are resized to $224\times 224$ and are applied to the network for classifying them as violent or non-violent. All the training video frames in a dataset are normalized to make their mean zero and variance unity.

\subsection{Datasets}

The performance of the proposed method is evaluated on three standard public datasets namely, Hockey Fight Dataset \cite{nievas2011violence}, Movies Dataset \cite{nievas2011violence} and Violent-Flows Crowd Violence Dataset \cite{Hassneretal:SISM12}. They contain videos captured using mobile phones, CCTV cameras and high resolution video cameras.\\
\textbf{Hockey Fight Dataset:} Hockey fight dataset is created by collecting videos of ice hockey matches and contains 500 fighting and non-fighting videos. Almost all the videos in the dataset have a similar background and subjects (humans). 20 frames from each video are used as inputs to the network.\\
\textbf{Movies Dataset:} This dataset consists of fight sequences collected from movies. The non-fight sequences are collected from other publicly available action recognition datasets. The dataset is made up of 100 fight and 100 non-fight videos. As opposed to the hockey fight dataset, the videos of the movies dataset is substantially different in its content. 10 frames from each video are used as inputs to the network.\\
\textbf{Violent-Flows Dataset:} This is a crowd violence dataset as the number of people taking part in the violent events are very large. Most of the videos present in this dataset are collected from violent events taking place during football matches. There are 246 videos in this dataset. 20 frames from each video are used as inputs to the network.

\subsection{Results and Discussions}

Performance evaluation is done using 5-folds cross validation scheme, which is the technique followed in existing literature. The model architecture selection was done by evaluating the performance of the different models on the hockey fight dataset. The classification accuracies obtained for the two cases, video frames as input and difference of frames as input, is listed in table \ref{table2}. From the table, it can also be seen that using a network that is pre-trained on the ImageNet dataset (we used BVLC AlexNet from Caffe model zoo) results in better performance compared to using a network that is randomly initialized. In this way, we decided to use frame difference as the input and to use a pre-trained network in the model.
Table \ref{table1} gives the classification accuracy values obtained for the various datasets considered in this study and is compared against 10 state of the art techniques. From the table, it can be seen that the proposed method is able to better the results of the existing techniques in the case of hockey fights dataset and movies dataset. 

As mentioned earlier, this study considers aggressive behavior as violent. The biggest problem of considering this definition occurs in the case of sports. For instance, in the hockey dataset, the fight videos consists of players colliding against each other and hitting one another. So one easy way to detect violent scenes is to check if one player moves closer to another. But the non-violent videos also consist of players hugging each other or doing high fives as part of a celebration. It is highly likely that these videos could be mistaken as violent. But the proposed method is able to avoid this which suggests that it is capable of encoding motion of localized regions (motion of limbs, reaction of involved persons, etc.). However, in the case of violent-flows dataset, the proposed method is not able to best the previous state of the art technique (it came second in terms of accuracy). Analyzing the dataset, it is found that in most of the violent videos, only a small part of the crowd is found to be involved in aggressive behavior while a large part remained as spectators. This forces the network to mark such videos as non-violent since majority of the people present in it is found to behave normally. Further studies are required for devising techniques to alleviate this problem involved with crowd videos. One technique that can be considered is to divide the frame in to sub-regions and predict the output of the regions separately and mark the video as violent if any of the regions is outputted by the network as violent.

In order to compare the advantage of convLSTM over traditional LSTM, a different model that consists of LSTM is trained and tested on the hockey fights dataset. The new model consists of the AlexNet architecture followed by an LSTM RNN layer. The output of the last fully-connected layer (fc7) of AlexNet is applied as input to an LSTM with 1000 units. The rest of the architecture is similar to the one that uses convLSTM. The results obtained with this model and the number of trainable parameters associated with it are compared against the proposed model in table \ref{table3}. The table clearly shows the advantages of using convLSTM over LSTM and the capability of convLSTM in generating useful video representation. It is also worth mentioning that the number of parameters that are required to be optimized, in the case of convLSTM, is very much less compared to LSTM (9.6M vs 77.5M). This helps the network to generalize better without overfitting in the case of limited data. The proposed model is capable of processing 31 frames per second on an NVIDIA K40 GPU.

\begin{table}
	\centering
	\caption{Comparison between convLSTM and LSTM models in terms of classification accuracy obtained in the hockey fights dataset and number of parameters}
	\begin{tabular}{l c c}
		\hline
		\textbf{Model} & \textbf{Accuracy} & \textbf{No. of Parameters} \\ \hline \hline
		convLSTM & 97.1$\pm$0.55\% & 9.6M(9619544) \\
		LSTM & 94.6$\pm$1.19\% & 77.5M(77520072) \\ \hline
	\end{tabular}
	\label{table3}
\end{table}
\vskip -2mm
\section{Conclusions}

This work presents a novel end-to-end trainable deep neural network model for addressing the problem of violence detection in videos. The proposed model consists of a convolutional neural network (CNN) for frame level feature extraction followed by feature aggregation in the temporal domain using convolutional long short term memory (convLSTM). The proposed method is evaluated on three different datasets and resulted in improved performance compared to the state of the art methods. It is also shown that a network trained to model changes in frames (frame difference) performs better than a network trained using frames as inputs. A comparative study between the traditional fully-connected LSTM and ~convLSTM is also done and the results show that the convLSTM model is capable of generating a better video representation compared to LSTM with less number of parameters, thereby avoiding overfitting.


{\small
\bibliographystyle{ieee}
\bibliography{AVSS17-violence}
}

\end{document}